\documentclass[10pt,twocolumn,letterpaper]{article}

\usepackage{wacv}              

\usepackage{graphicx}
\usepackage{amsmath}
\usepackage{amssymb}
\usepackage{booktabs}
\usepackage[accsupp]{axessibility}  

\usepackage[pagebackref,breaklinks,colorlinks]{hyperref}

\usepackage[capitalize]{cleveref}
\crefname{section}{Sec.}{Secs.}
\Crefname{section}{Section}{Sections}
\Crefname{table}{Table}{Tables}
\crefname{table}{Tab.}{Tabs.}

\def\wacvPaperID{2525} 
\def\confName{WACV}
\def\confYear{2024}

\begin{document}

\title{MIVC: Multiple Instance Visual Component for Visual-Language Models}

\author{Wenyi Wu\\
Amazon\\
424 5th Ave, New York, NY\\
{\tt\small wenyiwu@amazon.com}
\and
Qi Li\\
Amazon\\
410 Terry Ave N, Seattle, WA\\
{\tt\small qlimz@amazon.com}
\and
Wenliang Zhong\\
The University of Texas at Arlington \\
500 UTA Boulevard, Arlington, TX\\
{\tt\small wxz9204@mavs.uta.edu}
\and
Junzhou Huang\\
The University of Texas at Arlington \\
500 UTA Boulevard, Arlington, TX\\
{\tt\small jzhuang@uta.edu}
}
\maketitle





\begin{abstract}
Vision-language models have been widely explored across a wide range of tasks and achieve satisfactory performance. However, it’s under-explored how to consolidate entity understanding through a varying number of images and to align it with the pre-trained language models for generative tasks. In this paper, we propose MIVC, a general multiple instance visual component to bridge the gap between various image inputs with off-the-shelf vision-language models by aggregating visual representations in a permutation-invariant fashion through a neural network. We show that MIVC could be plugged into the visual-language models to improve the model performance consistently on visual question answering, classification and captioning tasks on a public available e-commerce dataset with multiple images per product. Furthermore, we show that the component provides insight into the contribution of each image to the downstream tasks.
\end{abstract}

\section{Introduction}

In recent years, numerous efforts\cite{goyal2017making,zhang2016yin} have been made to integrate images and text with multimodal models that typically utilizes distinct encoders for different modalities of data (e.g., CNNs as visual encoders and RNNs as text encoders). These encoders are subsequently fused in a shared embedding space. More recently, with the evolution of Transformer architectures, several studies\cite{li2023blip,liu2023visual,zhu2023minigpt} have sought to unify vision and text using text and vision Transformers. These methods commonly combine information from images and text tokens, enabling collaborative attention mechanisms within the Transformer for enhanced information fusion.

Despite the remarkable achievements of these methods in various multimodal tasks such as Visual Question Answering (VQA)\cite{antol2015vqa} and Image Captioning\cite{rennie2017self,you2016image}, there is an evident limitation. They often assume that input images and text are paired, meaning one image corresponds to one piece of text. However, in practical scenarios, this assumption can be challenged, as not all tasks involve a one-to-one relationship between image and text. For instance, when presented with two images, we may seek a textual description highlighting their differences. In this context, there exists a one-to-two relationship between text and images. Furthermore, multiple images may correspond to a single piece of text, especially when describing a complex object. For example, in e-commerce platform, each product is displayed with different background, from different angles or focusing on local details to provide enriched information. These images are correlated and essentially representing the same entity and therefore, it's crucial to learn a consolidated entity representation consolidating all images that could be aligned with the pre-trained language models for general generative tasks.

Given that state-of-the-art (SOTA) multimodal models\cite{dai2023instructblip, li2023blip} are primarily pre-trained on the one-image-one-text paradigm, directly inputting multiple images with one piece of text is unfeasible. Consequently, existing approaches typically address this issue through two methods: (1) when processing image inputs, they concatenate multiple images into a single "concentrated" image, or (2) they employ multiple images' embeddings obtained via encoders as input, although this often requires fine-tuning to adapt the model to multiple visual embeddings. However, these methods are simplistic in their approach to fusing information from multiple images. How to more effectively integrate one-to-many or many-to-many of image-text data remains an open question.

In this paper, we tackle the challenge regarding how to consolidate information from multiple images and text within a visual-language model, particularly when using multiple images to describe a single object. This problem is crucial in e-commerce~\cite{collins2022abo}, where a product is typically represented by multiple images along with a textual description to comprehensively convey its attributes. Notably, the scenario of multiple images describing an object differs from traditional multi-view problems, where multiple images possess information about relative positions. In our case, multiple images are simply used to describe the same object without requiring strong assumptions among them. Inspired by the Multiple-Instance Learning (MIL) problem\cite{carbonneau2018multiple,maron1997framework} where each input contains the varying number of entities, forming a set referred to as a bag, we consider the input images as a bag as well and aim to learn a consolidated representation per bag.

\begin{figure*}[tbh]
\centering
  \includegraphics[width=0.9\linewidth]{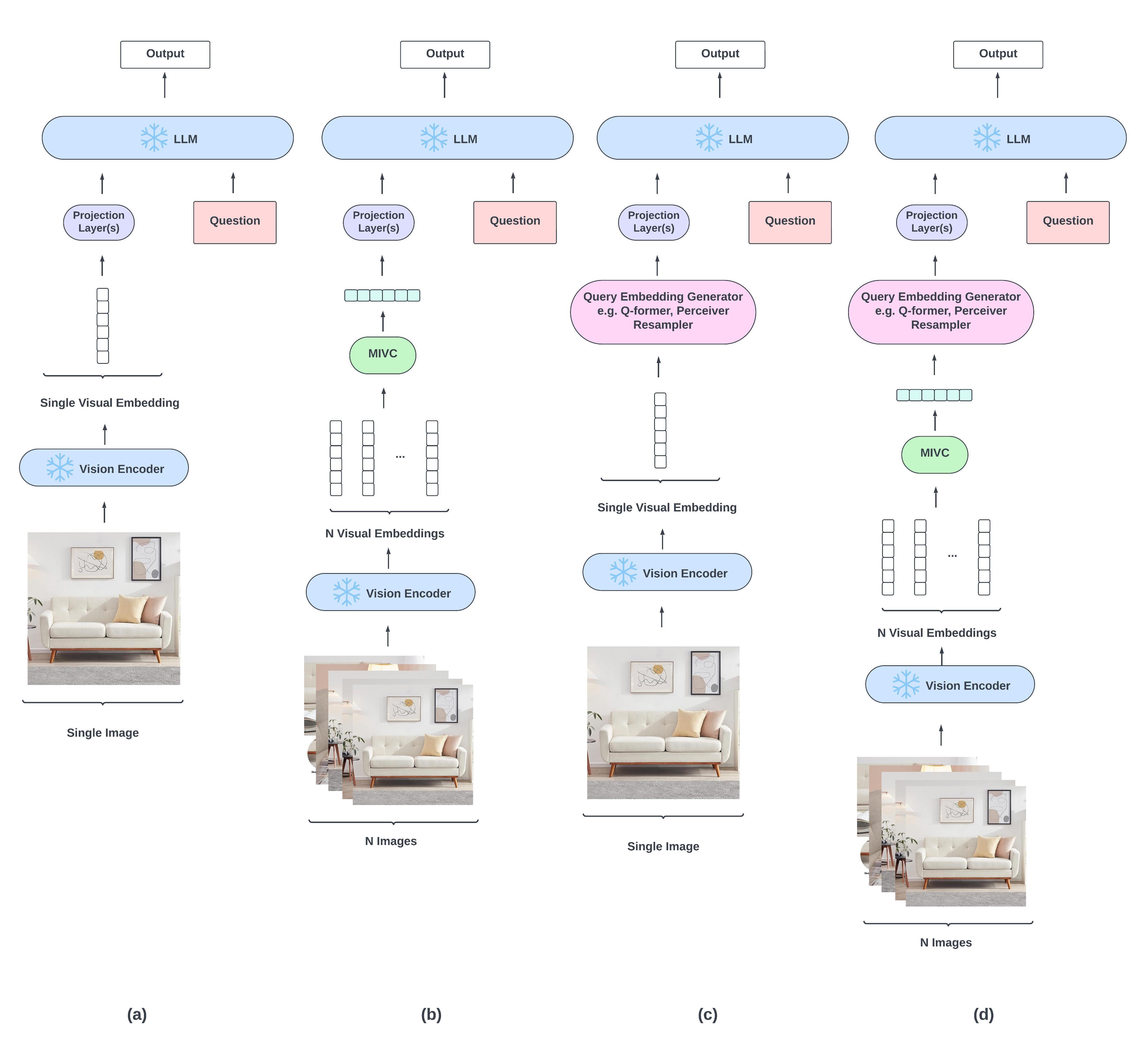}
   \caption{MIVC adaptation with off-the-shelf visual-language models. Both (a) and (b) utilize foundation vision encoder's output image patch embedding as visual representation. In contrast to (a), (b) is adapted with MIVC, which can take multiple images as input. Both (c) and (d) employ query embedding generator (e.g. Q-former \cite{li2023blip} or Perceiver Resampler \cite{alayrac2022flamingo}) after vision encoder for visual representation generation. In contrast to (c), (d) is adapted with MIVC which can takes multiple images as input. In this paper, we use T5-XXL as the language model and ViT as the vision encoder.}
\label{fig:arch}
\end{figure*}

Specifically, we leverage off-the-shelf vision encoders to convert each image into a representation and employ attention mechanisms to effectively combine multiple images within a bag through multiple instance learning. These combined embeddings are then used as input, alongside text, to the off-the-shelf language models for generative tasks. This approach not only allows us to accept multiple images as input but also identifies the most relevant images for the task through attention, thereby enhancing the model's performance and providing interpretability. In summary, our primary contributions include: (1) we introduce a groundbreaking Multiple Instance Learning (MIL) component, MIVC \ref{fig:arch}, in the realm of multimodal representation learning. Our novel framework enables the adaptive integration of multiple images with textual data, a critical advancement in handling complex multimodal information; (2) through our innovative MIL framework, we achieve a significant improvement in multimodal representation learning. This enhancement contributes to more effective fusion of information from diverse modalities, promising substantial benefits in various applications. We validate the effectiveness of the proposed method through extensive experimentation on the publicly available Amazon Berkeley Objects Dataset (ABO)~\cite{collins2022abo}. Our empirical results demonstrate its prowess in addressing real-world tasks, underlining its practical utility and robustness; and (3) providing insights into the contribution of each image to the generative tasks.

\section{Related Work}
\subsection{SOTA Visual-Language Models}
With the continuous evolution of deep learning, an increasing number of research endeavors have shifted their focus towards the fusion of different modalities to address more complex scenarios in the real world. For instance, in Visual Question Answering (VQA), users may pose questions to models based on a set of images, expecting answers. In image captioning, users provide a set of images, asking the model to generate descriptive text regarding the content of these images. In early works\cite{antol2015vqa,goyal2017making,zhang2016yin}, images and text were separately transformed into features using ResNet and LSTM, followed by concatenation before being fed into a prediction layer for inference.

In recent years, with the emergence of Transformers\cite{vaswani2017attention}, there has been a revolutionary shift in the universal network architecture in the field of Natural Language Processing (NLP). The utilization of the global attention mechanism within Transformers has become increasingly prevalent. Subsequently, exploration of Transformers in the field of Computer Vision (CV) has also taken flight. Vision Transformers (ViT)\cite{dosovitskiy2020image}, for example, employ the same Transformer architecture as in NLP but divide images into several patches to be treated as vision tokens as input. With extensive pre-training on large scale data, ViT has demonstrated superior performance to ResNet. Through Transformers, CV and NLP have achieved structural unification which enables multimodal model development. For example, phrase grounding \cite{gupta2020contrastive, wu2022catalog} aligns the visual signals with arbitrary caption words semantically, which extends the object detection task beyond the fixed list of categories in the label set.

More recently, with the rise of generative models\cite{brown2020language,raffel2020exploring,touvron2023llama}, the application of multimodal capabilities to generative tasks has become an open question. DALL-E\cite{ramesh2021zero}, for instance, embarked on a pretraining task where images were tokenized, enabled text-to-image generation. Subsequently, various vision-language models (VLMs) have been proposed to enhance the fusion of text and images. For example, BLIP2\cite{li2023blip} introduced the use of a Q-Former to align images more effectively with the input space of text. TCL\cite{yang2022vision} employed triplet contrastive learning to simultaneously learn from text and images. FROMAGe\cite{koh2023grounding} adopted a multitask approach to train a model for image captioning and image retrieval.

While these multimodal models have achieved substantial success across various tasks, they are predominantly built upon a crucial assumption - that a single piece of text pairs with a single image as input. However, in the real world, text and images may exhibit one-to-many or many-to-many relationships. How to effectively handle multimodal models in such scenarios remains an open question.

\subsection{Multiple Instance Learning}
Traditionally, Multiple Instance Learning (MIL)\cite{carbonneau2018multiple,maron1997framework} can be broadly categorized into two main types: (1) Bag-Level Prediction\cite{campanella2019clinical, feng2017deep, hou2016patch, kanavati2020weakly}: In this approach, bag-level predictions are directly derived from instance-level predictions. (2) Bag-Level Prediction with Feature Aggregation\cite{ilse2018attention,li2021dual,lu2021data,shao2021transmil}: Here, bag-level predictions are generated by aggregating the features of all instances. For the former, often, hard-crafted pooling operators such as mean pooling or max pooling are employed. However, in practical applications, these hard-crafted pooling operators often yield limited results. 

Aggregating instance features to form bag-level features typically leads to better outcomes but requires more complex pooling operations. Recent research has applied neural networks to the pooling process in MIL. For instance, MI-Net\cite{wang2018revisiting} utilizes a fully connected layer in MIL. Furthermore, AB-MIL\cite{ilse2018attention} employs attention during the pooling process, allowing for better weighting of different instances. Another category of methods attempts to consider the relationships between different instances using graph neural networks or capsule neural networks. More recently, DS-MIL\cite{li2021dual} employs attention not only to consider instance-to-instance relationships but also instance-to-bag relationships; DTFD-MIL\cite{zhang2022dtfd} incorporates the Grad-CAM mechanism into AB-MIL. While all these approaches focus on single modality, we adopt the effective attention mechanism proposed by AB-MIL to consolidate visual features in the visual-language models.

\section{Method}

\subsection{Architecture Overview}

By incorporating visual models with the capabilities of pre-trained Large Language Models (LLMs), multimodal LLMs have demonstrated dramatic improvements in various tasks, such as visual question answering (VQA), captioning, and etc. The majority of recent multimodal LLMs \cite{liu2023visual, dai2023instructblip, li2023blip} share a similar framework by utilizing separate vision and text towers to independently encode the two modalities first. The encoded single modality representations are then fused together, e.g. by projecting image representation via a single or multiple projection layers, or by directly concatenating, and then fed into LLMs. Depending on how image representation embedding is generated, it can be further categorized into two types: first, image patch embedding based vision tower in Figure \ref{fig:arch} (a), which is generally composed of a single visual foundation model encoder (e.g. ViT \cite{dosovitskiy2020image}) and utilizes the generated image patch embedding directly as visual representation; and second, image patch and query embedding based vision tower in Figure \ref{fig:arch} (c),  which sequentially combines a vision foundation model's encoder and a query embedding module (such as Q-former in BLIP-2 \cite{li2023blip} or the Perceiver Resampler as in Flamingo \cite{alayrac2022flamingo}).

One constraint to such a framework is the lack of capability to process multiple image inputs per request, when all images contribute and correspond to a single label. These multiple images, also referred to as multiple instances, typically carry complementary information; therefore they are more informative than a single instance for the corresponding task and shouldn't be ignored. Such applications are not rare in industry and other scientific areas, including utilizing multiple product images corresponding to a single product for e-commerce-related classification, caption generation, product information inference, synthesizing multiple X-ray images for medical diagnosis \cite{ilse2018attention}, and geological simulation from multiple underground mapping \cite{li2018unsupervised}, among others. To the best of our knowledge, all current visual-language models only consider a single image instance as input. Although it can be adapted to multiple image instances, this is largely achieved through customization at the input stage, either by taking only one single image as input or by concatenating multiple raw images into a single image. This results in either information loss, as multiple images' information is not efficiently synthesized, or a computational burden on LLM inference when dealing with one large-scaled concatenated image embedding. 

To address the accuracy and efficiency challenges, we propose MIVC, a general multiple-instance visual component that bridges the gap between multiple image inputs and any off-the-shelf Vision Language Models (VLMs). The proposed component can robustly handle both multiple image instances learning and single image instance learning. In particular, as illustrated in Figure \ref{fig:arch} (b, d) , compatible with any VLMs, we attach MIVC directly after the vision encoder. Multiple images are fed to the vision encoders to generate multiple visual representations via any VLMs' vision tower which retrieves the visual information from each individual image instance. The generated visual representations are then fed into MIVC to generate a single pooling image representation. This fused image representation not only retains essential information from multiple image instances but is also concise enough without introducing extra computational cost in the following LLM inference stage, where the pooling image representation is concatenated with text embedding and fed into the LLM for final inference. In this paper, to illustrate the effectiveness of MIVC, we use off-the-shelf pre-trained language model T5-XXL \cite{chung2022scaling} as the large language component and ViT \cite{dosovitskiy2020image} as the vision encoder. We compare the performance and computational complexity of the aforementioned natural alternatives with MIVC in the following sections.

\subsection{MIVC Methods}
\begin{figure}[htbh]
\centering
  \includegraphics[width=0.9\linewidth]{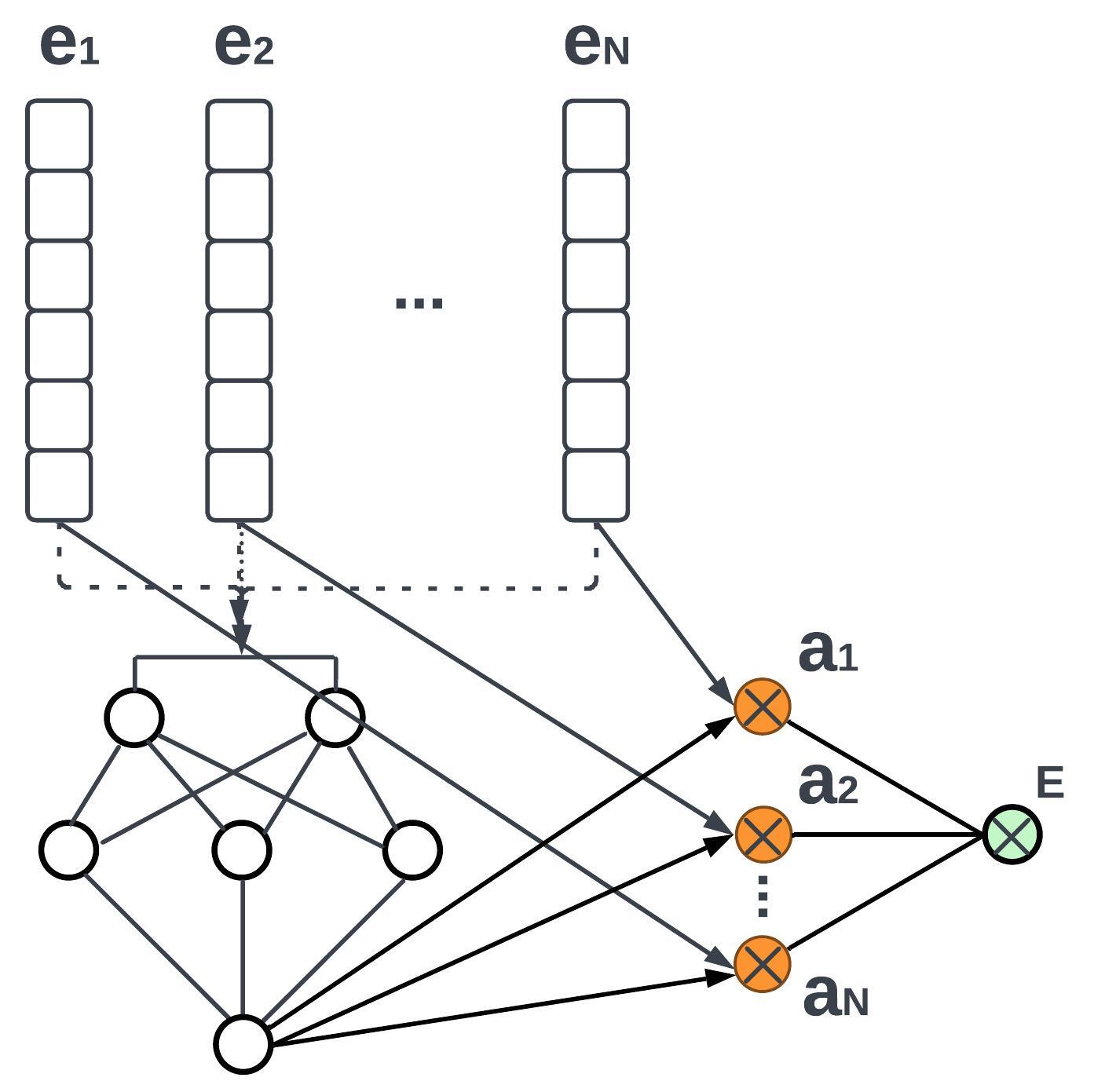}
   \caption{Detail structure of MIVC with attention pooling. The neural networks are trained to consolidate visual representations in a permutation-invariant fashion. The orange neurons represent the contribute of each image to the downstream generative tasks.}
\label{fig:pooling}
\end{figure}
\textbf{Problem Statement}
In the multiple instance learning problem, we have a bag of N instances, in any order, associated with a single label. In our problem setting, a single input data, $X = {I_1, I_2, ..., I_N, T, L}$, is composed of N images ($I_n \in \mathbb{R}^{L \times H \times C}$, where $n \in (1, N)$), along with a text prompt (T) and a text output (L). The value of $N$ varies for different input data. For a vision encoder $f$ in Vision Language Models (VLMs), we can generate N vision representations:
\begin{align}
    \forall_{n=1,...,N}: e_n = f(I_n) \ \ \ s.t.\ \  e_n \in \mathbb{R}^{M}
\end{align}
On top of that, we apply MIVC, our proposed multiple instance learning visual component, to generate a single pooling image representation:
\begin{align}
    E = MIVC(e_1,...,e_N)
\end{align}
In case of 2-dimensional vision representation, we first flatten them and then convert back to the original dimensions after MIVC.  We explore and evaluate four types of pooling strategies in the paper.

\textbf{Multiple Instance Pooling Strategy}

Following \cite{ilse2018attention}, we implemented four different embedding pooling strategies in MIVC :
\begin{itemize}
    \item Average pooling. Average operator across multiple image instances representation embeddings.
        \begin{align}
        E = \frac{1}{N} \sum_{n = 1}^{N} e_n
        \end{align} 

    \item Max pooling. Maximum operator across each dimension of multiple image instances representation embeddings.
        \begin{align}
            \forall_{n=1,...,N} : E_n=\max_{m=1,...,M}{e_{nm}} 
        \end{align}
    \item Attention pooling. We illustrate attention based pooling in the Figure \ref{fig:pooling}. It's a weighted average of multiple image instances representation embeddings.
        \begin{align}
           E =\sum_{n=1}^{N}\alpha_n e_n,
        \end{align}
        where:
        \begin{align}
            &\alpha_n = \frac{exp\{w^Ttanh(Ze_n^T)\}}{\sum_{j=1}^{N}exp\{w^Ttanh(Ze_j^T)\}} \\
            s.t. 
            &\sum_{n=1}^{N}\alpha_n = 1
        \end{align}
        
    in which $w \in \mathbb{R}^{K \times 1}$ and $Z \in \mathbb{R}^{K \times M}$ 
    \item Gated Attention pooling. The gated attention pooling introduces more non-linearity, and pooling more rich visual information across the multiple image instances.
        \begin{align}
           E =\sum_{n=1}^{N}\alpha_n e_n,
        \end{align}
        where:
        \begin{align}
            &\alpha_n = \frac{exp\{w^T(tanh(Ze_n^T) \otimes sigm(Ge_n^T))\}}{\sum_{j=1}^{N}exp\{w^T(tanh(Ze_j^T) \otimes sigm(Ge_j^T)\}} \\
            s.t. 
            &\sum_{n=1}^{N}\alpha_n = 1
        \end{align}
        in which $w \in \mathbb{R}^{K \times 1}$, $Z \in \mathbb{R}^{K \times M}$, $G \in \mathbb{R}^{K \times M}$, and $\otimes$ represents element-wise mulplication operator.
\end{itemize} 

\section{Complexity}
We measure the model's complexity by calculating the number of parameters associated with each pooling method, as detailed in Table \ref{tab:num_params}. From the table, it can be observed that both the average and max pooling methods introduce no additional trainable parameters, resulting in the same overall parameter count as the original BLIP2 model. In contrast, the Attention and Gated Attention methods introduce new parameters due to the inclusion of attention modules. However, the proportion of these additional parameters is minimal, accounting for only 0.7\% and 1.5\% of the total model parameters, respectively. Consequently, their supplementary computational overhead is negligible.

\begin{table}[!h]
    \centering
    \resizebox{0.48\textwidth}{!}{
        \begin{tabular}{c|c}   
            \hline
            \text{Models w/ pooling} & \text{\# params} \\
            \hline
            T5-XXL+ViT & 12.23B \\
            T5-XXL+ViT w/ avg or max & 12.23B \\
            T5-XXL+ViT w/ attn (extra params) & 12.32B (92.23M) \\
            T5-XXL+ViT w/ gated attn (extra params) & 12.42B (185.63M) \\
            \hline
        \end{tabular}
    }
    \caption{Model complexity in terms of the number of parameters. This directly impacts the inference efficiency.}
    \label{tab:num_params}
\end{table}

\section{Training}
For attention and gated attention based MIVC, it learns the neural network $w$,$Z$ and $G$ which determine how each image contributes to the downstream generative tasks. To make fair comparison between different pooling methods and vanilla alternatives, besides the zero-shot evaluation of the off-the-shelf models, we further fine-tune all models to report performance. We use product images and textual metadata to train the image-textual alignment layers and MIVC simultaneously using generative tasks. Both dataset and task details are presented in ~\ref{sec:data} and ~\ref{sec:task}. We freeze the visual encoder and the language model during our training procedure.  

\section{Experiment and Analysis}
\subsection{Datasets}
\label{sec:data}
\begin{figure*}
\centering
\includegraphics[width=0.9\linewidth]{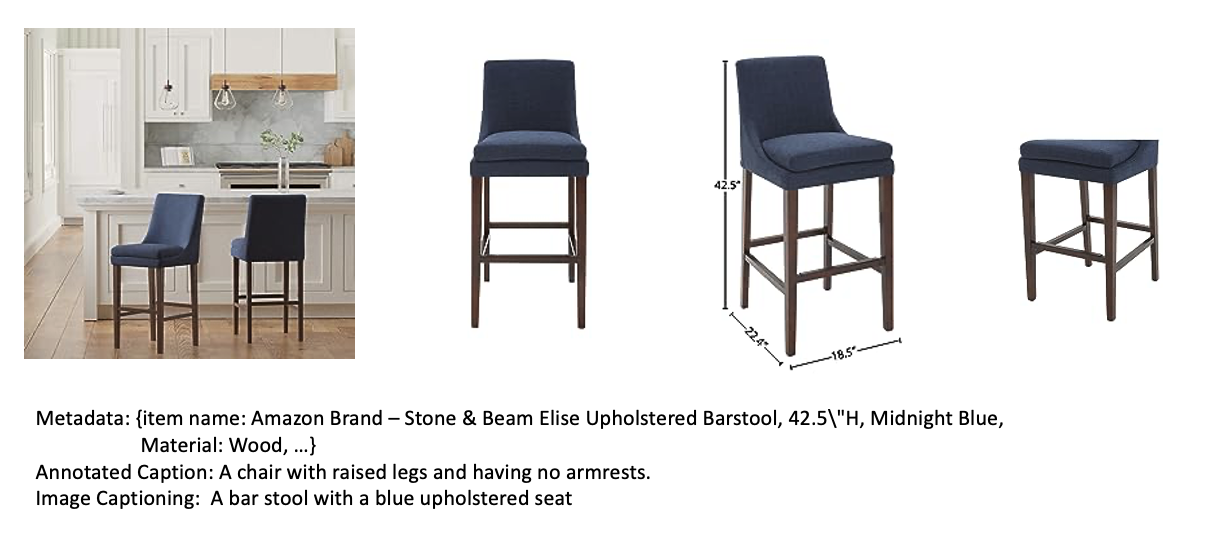}
   \caption{Data illustration. The e-commerce product contains one attractive main images (the leftmost image), several detailed images and textual metadata. Instead of item name displayed in e-commerce website, we use manually annotated image captions to measure captioning performance.}
\label{fig:data_exp}
\end{figure*}

The products in the e-commerce website are presented by one main image and multiple images from different views or with zoomed in images to display details, such as patterns, flavours and etc. We leverage ABO dataset ~\cite{collins2022abo} with 147,702 products which contains multiple images and textual metadata as shown in the e-commerce website. The number of images per product ranges widely from 2 to 21 and images could be with different background, angles and focal lengths. The main image is commonly attractive by putting the product into a live scene which makes it hard to focus on the correct entity or detail region for generative tasks requiring detailed vision signals. On the contrary, the subsequent images with white background or of details benefit generative models with fine-grained visual representations. The textual metadata mainly contains one short descriptive sentence to be displayed as the title in the e-commerce product page and seven product attributes, e.g., color, pattern, style and etc. We illustrate one product example and the generated caption using the MIVC-BLIP2 model in Figure \ref{fig:data_exp}.   

\subsection{Tasks}
In total, we have 3 types of tasks. We split each of them into $80\%$ training set and $20\%$ evaluation set. All training sets are mixed and the model are trained with the unified generative tasks \cite{radford2019language} despite they are evaluated differently.

\label{sec:task}
\textbf{Categorization} Product categorization is an important task for e-commerce which benefits search and recommendation experience. In ABO, there exist hundreds of categories with a long tail distribution. We keep 10 high frequency and representative categories to test our MICV including furniture, shoes and etc. We intent to keep categories that are not trivial to distinguish like chair and sofa. We form this as a multi-choice visual question answering task with 10 options and feed the MIVC-BLIP2 model with multiple product images, the categorization question and 10 candidate categories. We compare the performance using accuracy, macro average precision, and recall across 10 categories. 

\textbf{Product Information Inference} Besides the product categorization, we further look into detail product metadata from seven attributes: style, color, finish type, pattern, fabric type, material and shape. Following the science QA ~\cite{lu2022learn} format, we frame them into multi-choice questions and generate answer based on images and the prompt. Because the metadata is not as clean as scientific question answering dataset, for example, $100\%$ cotton and pure cotton both exist, we clean the dataset and apply regex to only keep alphabetic characters. We further keep top frequent values per attribute to be used as multi-choice answers. The number of values ranges from 3 to 5 across different attributes. For example, we have solid, textual, floral, geometric and striped in pattern. We compare the performance using accuracy, macro average precision, and recall across choices for each attribute and report the aggregated performance across all attributes. 

\textbf{Image Captioning} We use the general image captioning prompts \cite{dai2023instructblip} to ask the model to generate a short descriptive sentence of the product using multiple images. Instead of using original product titles, which are often less descriptive, we leverage the manually annotated captions \cite{luo2023scalable} on the same dataset as the reference captions to measure the quality of our generated captions. It's not trivial to evaluate the quality of the generated captions because each product could be described from very different perspectives. We illustrate how the generated title could be different from the reference title but still very relevant to the image content in Figure \ref{fig:data_exp}. Therefore, we feed the generated captions, annotated captions and the main product image to the pre-trained CLIP model \cite{radford2021learning} to retrieve text given image. We report the text retrieval top 1 recall as our metrics, which is the proportion of samples whose generated caption has the highest similarity with the image compared to all annotated captions.

\subsection{Benchmark Models}
\textbf{Single Image} We evaluate the T5-XXL and ViT on the aforementioned data and tasks using only the first image in zero-shot fashion.

\textbf{Concatenated Image} Another vanilla approach to infer visual signals from multiple images is to concatenate all images together, as being illustrated in \cite{dai2023instructblip,liu2023visual}. Because the number of images could be as many as 21, horizontal concatenation will lead to aspect ratio challenge. Therefore, we concatenate images in a square grid such that 4 images are concatenated in a 2 by 2 grid, 5 to 9 images are concatenated in a 3 by 3 grid with blank image fill-in and so on and so forth. We evaluate the BLIP2 on the aforementioned data and tasks using the concatenated image in zero-shot fashion.
\begin{table}[htbp]
  \begin{center}
    {\small{
 \begin{tabular}{llllll}
    \toprule
        Pooling & Accuracy & Precision & Recall \\
    \midrule
    Single (zs)  & 97.1\% & 97.1\% & 97.1\%  \\ 
    Concat (zs) & 97.5\% & 97.6\% & 97.5\%  \\ 
    Single  & 97.2\% & 97.3\% & 97.2\%  \\ 
    Concat  & 97.8\% & 97.8\% & 97.8\%  \\ 
    MIVC-Avg  & 96.9\% & 97.0\% & 96.9\%  \\ 
    MIVC-Max  & 94.7\% & 94.7\% & 94.7\% \\ 
    MIVC-Attn  & 97.9\% & 97.9\% & 97.9\%  \\ 
    MIVC-gated & 97.4\% & 97.4\% & 97.4\%  \\ 
    \bottomrule
    \end{tabular}
}}
\end{center}
\caption{Categorization performance comparison. We illustrate the effectiveness of MIVC with T5-XXL and ViT as language model and the vision encoder, respectively. We first evaluate them in zero-shot (zs) fashion and then fine-tune the model with and without various MIVC pooling method. The reported precision, recall and f1-score are macro average across 10 categories.}
\label{tab:pt-cls}
\end{table}

\subsection{Results and Analysis}
We summarize the performance of three tasks in the following tables. From the table \ref{tab:cap}, the results show that the proposed MIVC with attention pooling outperforms all benchmarks on the 10 categories classification task. From the table \ref{tab:vqa}, we observe that the MIVC with attention pooling outperforms the benchmark methods by selecting the most accurate options from candidates in the product inference task. It improves the performance the most by $9\%$ accuracy, compared to the single image benchmark. It aligns with our conjecture that including additional images would benefit generative tasks with fine-grained visual information. From the table \ref{tab:cap}, we show that with MIVC, the general visual-language can generate high quality comprehensive titles.

\begin{table}[htbp]
  \begin{center}
    {\small{
 \begin{tabular}{llll}
     \toprule
         Pooling & Accuracy& Precision & Recall \\ 
        \midrule
        Single (zs) & 64.5\% & 65.3\% & 63.0\% \\  
        Concat (zs) & 65.8\% & 68.9\% & 65.5\% \\  
        Single & 62.9\% & 62.8\% & 62.9\% \\  
        Concat & 66.0\% & 68.0\% & 65.0\% \\  
        MIVC-Avg  & 64.7\% & 65.7\% & 63.7\% \\  
        MIVC-Max & 65.8\% & 66.7\% & 64.7\% \\  
        MIVC-Attn & 67.4\% & 72.7\% & 70.1\% \\  
        MIVC-gated & 66.9\% & 68.5\% & 65.1\% \\  
        \bottomrule
    \end{tabular}
}}
\end{center}
\caption{Product attribute inference performance comparison. We illustrate the effectiveness of MIVC with T5-XXL and ViT as language model and the vision encoder, respectively. We first evaluate them in zero-shot (zs) fashion and then fine-tune the model with and without various MIVC pooling. The reported precision, recall and f1-score are first macro average across the number of options per attribute and then simple averaged across tasks.}
\label{tab:vqa}
\end{table}
\begin{figure*}[ht]
\centering
\includegraphics[width=0.8\linewidth]{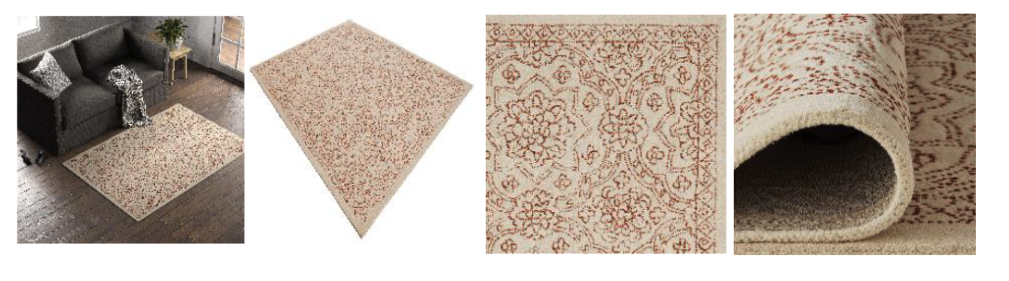}
   \caption{Interpretability: four images of the same product are fed to the VQA task to identify the pattern of the product. The weights from left to right are [0.24, 0.05, 0.65, 0.06] indicating that the region image contributes more to the pattern recognition.}
\label{fig:weights}
\end{figure*}

\begin{table}[htbp]
  \begin{center}
    {\small{
\begin{tabular}{ll}
    \toprule
        Image Pooling & Text Retrieval R@1\\   
        \midrule
        Single (zs) & 79.1\% \\   
        Concat (zs) & 77.4\% \\   
        Single & 76.0\% \\   
        Concat & 76.8\% \\   
        MIVC-Avg  & 76.7\% \\   
        MIVC-Max & 77.6\% \\   
        MIVC-Attn & 81.7\% \\   
        MIVC-gated & 80.2\% \\   
        \bottomrule
    \end{tabular}
}}
\end{center}
\caption{Image captioning performance evaluation. We illustrate the effectiveness of MIVC with T5-XXL and ViT as language model and the vision encoder, respectively. We first evaluate them in zero-shot (zs) fashion and then fine-tune the model with and without various MIVC pooling. We report the top 1 recall of retrieving the generated captions among manual annotated captions given the image.}
\label{tab:cap}
\end{table}

\subsection{Ablation: Concatenation}
Besides concatenating raw images, another natural alternative is to concatenate the image representations generated from the vision encoder and then project the concatenated image embedding to lower dimension space. For the BLIP series model using Q-former, the image representation is $257 \times 1408$ dimension. In order to project $N$ concatenated image embeddings back to the same dimension, it results $N \times 97B$ parameters where $N$ is the maximum number of images per product , i.e. 21. To make the number of parameters under control, we first limit the maximum number of input images to $6$ and then map image representations to a lower dimensional $2048$ before mapping to the proper dimension to the Q-former. We ends up with 4B parameters. To understand the performance loss caused by input image limitation and dimension reduction, we compare the performance of this approach against the above mentioned models on the pattern attribute recognition task, which is one of the above mentioned VQA questions that require fine-grained image signals. The results in Table \ref{tab:ablation} shows that after training the projection layers, the embedding concatenation performs worse than the rest models.

\begin{table}[htbp]
  \begin{center}
    {\small{
\begin{tabular}{llll}
     \toprule
        Pooling & Accuracy & Precision & Recall \\  
        \midrule
        single image & 51.8\% & 58.5\% & 54.6\% \\  
        concate image & 53.4\% & 59.5\% & 55.4\% \\  
        concate embed & 49.6\% & 51.3\% & 49.8\% \\  
        MIVC-Avg  & 51.2\% & 53.1\% & 51.6\% \\  
        MIVC-Max & 53.5\% & 54.8\% & 53.9\% \\  
        MIVC-Attn & 68.4\% & 71.0\% & 69.9\% \\  
        MIVC-gated & 64.3\% & 66.4\% & 63.7\% \\
        \bottomrule
    \end{tabular}
}}
\end{center}
\caption{Ablation regarding the embedding concatenation. After training, the model with concatenated image embeddings perform worse than the rest model on one of the VQA tasks: product pattern recognition.}
\label{tab:ablation}
\end{table}

\subsection{Interpretability}

The attention pooling in the MIVC generates a weighted average of visual representations for a bag of input images where the weights are parameterized by the neural network that are learned during training. These weights provide insights on which image contributes the most to the downstream tasks, as illustrated in \ref{fig:weights}. In the example, the rug contains 4 images, where the first one is the rug in the live scene background and the third is enlarged local pattern details. The attention based pooling method learns to mainly focus on pattern details in the third image to infer the generative task. The lower weights of the second the the last images may cause by the fact that the second image is distorted and vague while the last image is too detail to contain useful information. The first image may provide additional context or usage information of the product that could be learned from the live scene background.

\section{Conclusion and Future Work}

In this paper, we propose MIVC, a multiple instance visual component to address the challenge where the visual representation of one entity should be inferred from multiple images. We show that MIVC outperforms the vanilla alternatives on the e-comemrce dataset where each product is presented by multiple images. We also explore various approaches to pool the image representations. This component is compatible with a wide range of vision-language models besides Flan-T5 model and Qformer used in this paper, which could be explored in the future. The attention-based pooling could be further improved by cross modality attention which could be explored in the future.


{\small
\bibliographystyle{ieee_fullname}
\bibliography{egbib}
}

\end{document}